\documentclass{article}
\usepackage{spconf,amsmath,amssymb,graphicx,booktabs,array,xcolor,url,hyperref,iftex,needspace,placeins,flushend}

\usepackage{color}

\hypersetup{hidelinks}
\ifXeTeX
  \usepackage{fontspec}
  \IfFontExistsTF{TeX Gyre Termes}
    {\setmainfont{TeX Gyre Termes}}
    {\IfFontExistsTF{Times New Roman}
      {\setmainfont{Times New Roman}}
      {\setmainfont{Latin Modern Roman}}}
\fi

\newcommand{\method}{UACR}
\newcommand{\thitA}{\mathrm{TypeHit\mbox{-}A}}
\newcommand{\thitP}{\mathrm{TypeHit\mbox{-}P}}

\newcommand{\rate}{\mathrm{Rate}}
\newcommand{\qwenNine}{Qwen3.5-9B}
\newcommand{\qwenSeven}{Qwen2.5-VL-7B-Instruct}
\newcommand{\unadapted}{Unadapted MLLM}

\title{UACR: UNCERTAINTY-AWARE COLLABORATIVE ROUTING FOR EXPLAINABLE INDUSTRIAL ANOMALY INSPECTION}
\name{Yiwei Li$^{1}$ \qquad Qiyang Chen$^{1}$ \qquad Zheming Yang$^{2,3}$\sthanks{\fontsize{9}{10.5}\selectfont Corresponding author}}
\address{ $^{1}$The Chinese University of Hong Kong, Shenzhen, China\\
$^{2}$Institute of Computing Technology, Chinese Academy of Sciences, Beijing, China\\
$^{3}$University of Chinese Academy of Sciences, Beijing, China
}

\begin{document}
\ninept
\maketitle

\begin{abstract}

Industrial anomaly inspection requires not only accurate anomaly discrimination but also trustworthy explanations for operators, while satisfying the strict latency constraints of online inspection. Existing methods, however, typically apply a fixed computational path to all samples, making it difficult to strike a good balance among detection accuracy, inference efficiency, and interpretability. To address this issue, we propose UACR, an uncertainty-aware collaborative routing framework for explainable industrial anomaly inspection. UACR aims to deliver reliable anomaly decisions together with operator-facing explanations under the online inspection budget, while adaptively allocating computation so that ambiguous cases receive deeper scrutiny. The framework consists of two stages. Stage I, Offline Training and Routing Hyperparameter Search, learns the anomaly inspection modules and calibrates the uncertainty estimator that drives routing. Stage II, Uncertainty-Aware Adaptive Routing, collaboratively dispatches each input across inspection paths according to its estimated uncertainty. Experimental results on multiple public datasets show that UACR achieves over 5.16\% improvement in accuracy compared to the baseline method, while reducing latency by over 31.5\%. It also enhances the interpretability of industrial detection by over 15\%.

\end{abstract}

\begin{keywords}
industrial anomaly detection, multimodal LLM, selective
routing, explainable inspection
\end{keywords}

\section{Introduction}
\label{sec:intro}

\begin{figure}[!t]
  \centering
  \includegraphics[width=\columnwidth]{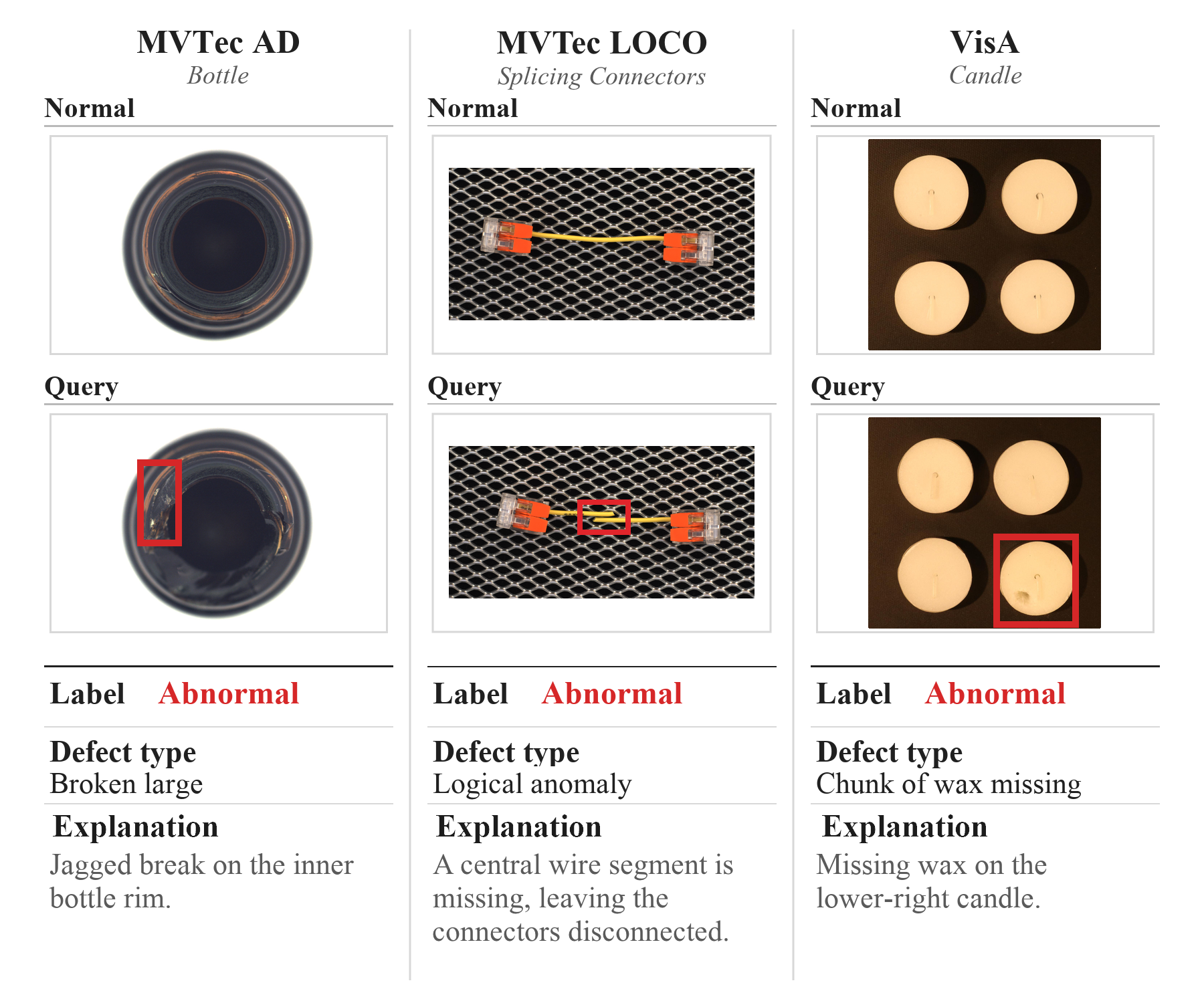}
  \caption{Held-out anomalies from MVTec AD, MVTec LOCO, and VisA. Each column
  pairs a same-category normal reference with an abnormal query. Red boxes
  mark the visible defect. The structured record reports the final label,
  defect type, and explanation.}
  \label{fig:examples}
\end{figure}

Reliable industrial anomaly detection is central to manufacturing quality
assurance because defects may be rare, visually subtle, and costly to annotate,
while production lines require timely and dependable decisions \cite{hu2026adaptive232323}. Accurate,
efficient, and actionable inspection is therefore essential for preventing
defective products from propagating through manufacturing and deployment \cite{hu2026aivd252525}. Lightweight screening models commonly operate in a cold-start regime in which
nominal images are available, but representative examples of every defect type
are not. Modern visual anomaly detection (VAD) methods model nominal feature
distributions, reconstruction behavior, or teacher--student discrepancies
\cite{defard2021padim,zavrtanik2021draem,roth2022patchcore,deng2022rd4ad,batzner2024efficientad}.
Compact variants make millisecond-scale local screening practical\cite{wang2023multimodal272727}. However,
category-dependent score scales and ambiguous samples near a decision boundary
can make a hard local decision unreliable, as shown in Fig.~\ref{fig:examples}. Moreover, an anomaly score or
evidence map does not by itself state the defect type or explain which visible
cue supports the decision, both of which are important for operator action.

Multimodal large language models (MLLMs) offer a semantic route beyond score-only inspection\cite{yang2025moa242424}. Language-guided VAD
methods such as WinCLIP and AnomalyCLIP exploit vision--language priors
\cite{jeong2023winclip,zhou2024anomalyclip}, while AnomalyGPT and MMAD show that
industrial anomalies can be queried and described with MLLMs
\cite{gu2024anomalygpt,jiang2025mmad}. Their richer reasoning comes with much
higher per-sample computation and latency than lightweight local screening, so
applying an MLLM to every production image is inefficient. A practical system
need instead decide which samples warrant semantic verification\cite{yang2026clap222222}. Selective prediction and deferral methods formalize accuracy--cost trade-offs
by rejecting difficult inputs
\cite{geifman2019selectivenet,mozannar2020defer}. The scene-aware enhanced
edge--cloud framework applies a related idea to industrial inspection by
routing according to scene complexity \cite{tian2025saec}. Scene complexity,
however, need not reflect the uncertainty of the deployed anomaly detector: a
visually simple image may lie close to its decision boundary, whereas a
cluttered but nominal image may still be easy. Existing routing therefore does
not fully reconcile local reliability, MLLM invocation overhead, and the need to
attach defect semantics to every local alarm.

To address the above problem, we introduce uncertainty-aware collaborative routing (\method{}), a collaborative inspection framework coupling a lightweight
anomaly detector with an MLLM verifier. Its routing signal is
tied to the category-specific decision boundary of the local detector. The
framework retains fast local handling for confident normal samples and invokes
a reference-conditioned semantic verifier wherever detector uncertainty or an
alarm makes additional reasoning valuable. Our contributions are as follows:
\begin{itemize}
  \setlength{\itemsep}{1pt}
  \setlength{\parsep}{0pt}
  \setlength{\parskip}{0pt}
  \item We propose UACR, an uncertainty-aware collaborative routing framework for explainable industrial anomaly inspection, which delivers reliable anomaly decisions and operator-facing explanations under strict online latency constraints by adaptively allocating computation to ambiguous cases.
  
  \item We develop a two-stage scheme that first learns the inspection modules and calibrates the uncertainty estimator offline, and then performs uncertainty-aware adaptive routing that collaboratively dispatches each input across inspection paths according to its estimated uncertainty.
  
  \item Experimental results on multiple public datasets show that UACR achieves over 5.16\% improvement in accuracy compared to the baseline method, while reducing latency by over 31.5\%. It also enhances the interpretability of industrial detection by over 15\%.
  
\end{itemize}

\section{Related Work}
\label{sec:related}

PaDiM models patch distributions in pretrained feature space \cite{defard2021padim}. PatchCore retains representative nominal patch features in a memory bank \cite{roth2022patchcore}. Reverse distillation increases teacher–student representation discrepancy on anomalies \cite{deng2022rd4ad}. SimpleNet learns a target-domain adapter and feature discriminator \cite{liu2023simplenet}. Unified multi-class models further seek to share representations across product categories \cite{you2022uniad}. These approaches are strong at score-based recognition and localization, but do not natively produce operator-facing explanations.

AnomalyGPT adapts a vision–language model with synthetic anomaly–text pairs \cite{gu2024anomalygpt}. MMAD exposes persistent weaknesses of general MLLMs on fine-grained industrial anomaly reasoning \cite{jiang2025mmad}. Reject-option classifiers optimize prediction quality under a coverage constraint \cite{geifman2019selectivenet}. Learning-to-defer methods further balance a model against a downstream expert \cite{mozannar2020defer}. SAEC uses image complexity to choose local or cloud inspection \cite{tian2025saec}. Although such models can generate explanations, they offer no explicit mechanism for quantifying explanation quality. They struggle to satisfy the strict latency and verifiability requirements of online industrial inspection.

\begin{figure}[t]
  \centering
  \includegraphics[width=\columnwidth]{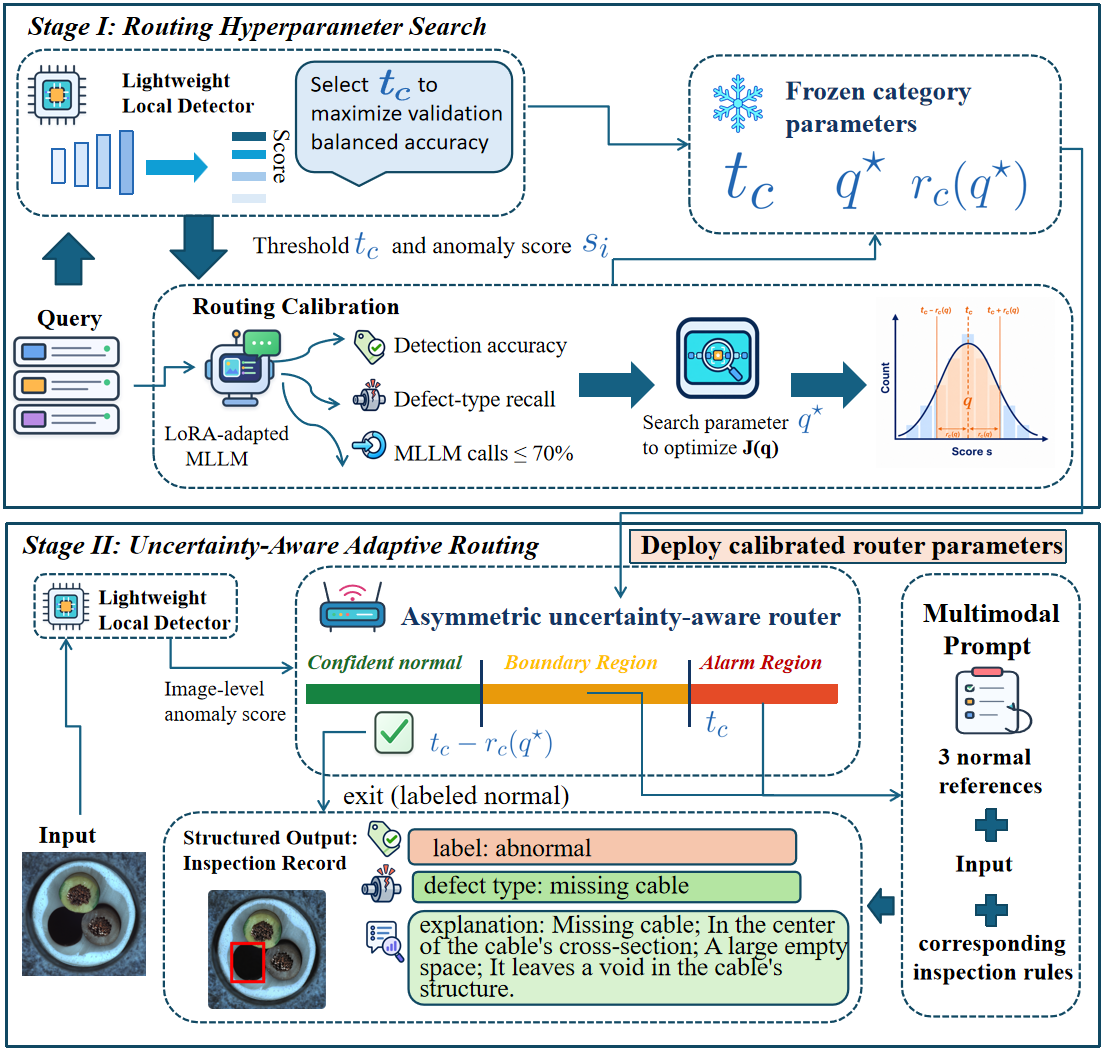}
  \caption{Overview of \method{}. Stage~I calibrates routing on the validation
  set using the already-trained local detector and LoRA-adapted verifier.
  Category-wise scores determine $t_c$. Full-cascade validation then selects
  $q^\star$ under the MLLM-call budget, yielding $r_c(q^\star)$. The routing
  configuration, references, and rules are frozen before evaluation. In
  Stage~II, confidently normal samples exit locally, while boundary cases and
  alarms are verified to produce a structured label, defect type, and
  explanation.}
  \label{fig:method}
\end{figure}

\section{Method}
\label{sec:method}
We propose UACR, an uncertainty-aware collaborative routing framework for explainable industrial anomaly inspection, as illustrated in Fig.~\ref{fig:method}. Its objective is to deliver reliable anomaly decisions together with operator-facing explanations, while adaptively allocating computation so that ambiguous cases receive deeper scrutiny without violating the online inspection budget. UACR consists of two stages. Stage I: \textit{Offline Training and Routing Hyperparameter Search}, learns the anomaly inspection modules and calibrates the uncertainty estimator that drives routing. Stage II: \textit{Uncertainty-Aware Adaptive Routing}, collaboratively dispatches each input across inspection paths according to its estimated uncertainty.

\subsection{Offline Training and Routing Hyperparameter Search}
\label{sec:offline}

We separate component-training data $\mathcal T$, labeled calibration data
$\mathcal V$, and evaluation queries $\mathcal E$. The local detector is
trained with its native anomaly-detection objective. A nominal-only detector
uses only normal images from $\mathcal T$. The semantic verifier is adapted
on available inspection records, conditioned on nominal references and
inspection rules. A field-weighted autoregressive objective is
\begin{equation}
 \mathcal L_{\mathrm{ver}}=
 \lambda_y\mathcal L_{\mathrm{label}}+
 \lambda_d\mathcal L_{\mathrm{type}}+
 \lambda_e\mathcal L_{\mathrm{explanation}},
 \label{eq:loss}
\end{equation}
where each term is token cross-entropy for the corresponding output field.
The positive weights control the relative supervision of the decision and
its semantics. Normal records use a normal-type token and a brief normality
statement. The choice of adaptation mechanism, reference count, and numerical
training settings belongs to the implementation, not the routing rule.

After training, both components are frozen. Score scales vary with texture,
geometry, and nominal variability, so calibration is performed within each
category rather than with a global numerical threshold. For validation subset
$\mathcal V_c$, we select
\begin{equation}
 t_c=\arg\max_{t\in\mathcal S_c}
 \tfrac{1}{2}\big(R_{\mathrm n}(t)+R_{\mathrm a}(t)\big),
 \label{eq:threshold}
\end{equation}
where $\mathcal S_c$ is the set of candidate score thresholds and
$R_{\mathrm n},R_{\mathrm a}$ denote normal and abnormal recall.
Balanced accuracy gives equal importance to both classes. The margin
$d_i=|f_{\ell}(x_i)-t_c|$ then defines an uncertainty radius:
\begin{equation}
 r_c(q)=\operatorname{Quantile}_{q}
 \big(\{d_i:x_i\in\mathcal V_c\}\big).
 \label{eq:radius}
\end{equation}
Here $q$ controls the width of a boundary interval in the category's own score
distribution. It is neither a calibrated error probability nor the eventual
MLLM invocation rate. For each candidate $q\in\mathcal Q$, we evaluate the complete cascade on
$\mathcal V$ using the gate defined in Eq.~\eqref{eq:gate}. Selection balances
three quantities:
\begin{equation}
 \begin{aligned}
 J(q)&=a\,\mathrm{Acc}(q)+b\,\thitA(q)-c_{\mathrm r}\,\rate(q),\\
 q^\star&=\arg\max_{\substack{q\in\mathcal Q\\\rate(q)\leq B}}J(q),
 \end{aligned}
 \label{eq:objective}
\end{equation}
where $a,b,c_{\mathrm r}>0$ are shared trade-off weights, $B$ is the invocation
budget, and all metrics are computed on validation data. TypeHit-A measures
defect-type recall, defined in Section~\ref{sec:experiment}. The same objective
and search protocol are used across tasks, while the selected operating point
may differ with the category mixture and verifier. Candidate policies can
reuse cached detector scores and verifier outputs, avoiding repeated semantic
inference during the search. We then freeze $t_c$, $q^\star$,
$r_c(q^\star)$, reference assignments, and inspection rules. Evaluation labels
are not inputs to this calibration procedure.

\subsection{Uncertainty-Aware Adaptive Routing}
\label{sec:online}

A cheap local
detector first screens every query, and only queries whose local evidence is
deemed unreliable are escalated to a reference-conditioned verifier that
returns defect semantics and an explanation. Let $s_i$ denote the local
anomaly score and $t_c$ the operating threshold calibrated offline. Rather
than committing to a hard decision at $t_c$, we define the routing indicator
from the \emph{uncertainty} of this decision:
\begin{equation}
 g_i=\mathbf{1}\!\left\{s_i\geq t_c\ \lor\
 |s_i-t_c|\leq r_c(q^\star)\right\},
 \label{eq:gate}
\end{equation}
where $r_c(q^\star)$ is the $q^\star$ quantile of the validation margin
$|s-t_c|$ over all validation samples in category $c$, so that $q^\star$ explicitly controls the
tolerated decision uncertainty rather than a raw score offset.
Equation~\eqref{eq:gate} encodes two escalation criteria: all local alarms
($s_i\geq t_c$) are escalated so that an abnormal decision is verified and
accompanied by defect semantics, and boundary cases
($|s_i-t_c|\leq r_c$) are escalated so that possible misses on the nominal
side can be re-examined. Only $s_i<t_c-r_c(q^\star)$ yields a local normal
exit. The policy is thus equivalent to a one-sided escalation threshold whose
width is set by an uncertainty quantile---which also clarifies why a small
$q^\star$ need not reduce the invocation rate, since the alarm region is
always included.

For an escalated sample, the verifier jointly examines the query and its
nominal references under the frozen inspection rules. It may confirm a local
alarm or correct either a false alarm or a missed anomaly, and its decision
is not forced to agree with the local score. The final record is
\begin{equation}
 (\hat y_i,\hat d_i,\hat e_i)=
 \begin{cases}
 F_{\theta}(x_i,\mathcal R_c,\mathcal P_c),&g_i=1,\\
 (\mathrm{normal},\mathrm{good},e_{\mathrm n}),&g_i=0,
 \end{cases}
 \label{eq:output}
\end{equation}
Here, $e_{\mathrm n}$ denotes a fixed local-normal status rather than a
generated defect explanation. The same field definitions are shared by both
paths. Therefore, downstream consumers observe a single output schema regardless of
route. The gate allocates computation and the verifier supplies semantic
judgment, and no training or parameter search is performed during online
inference. The formulation applies to any score-producing local detector
paired with a reference-conditioned structured verifier, provided that the
local scores are calibrated on representative validation data and that the
nominal references and inspection rules are appropriate for the target category.

\section{Experiments and Results}
\label{sec:experiment}

\subsection{Experimental setup}
\label{sec:setup}
\textbf{Implementation Details.}
Experiments were conducted using two NVIDIA H100 GPUs under Ubuntu 22.04.4 LTS and Python 3.10.20. The local scorer is EfficientAD \cite{batzner2024efficientad}, a lightweight
teacher--student anomaly detector with an autoencoder branch. The semantic backbones are \qwenNine{} (9B) \cite{qwen35modelcard} and \qwenSeven{} (7B) \cite{bai2025qwen25vl}. Each query is paired with three same-category nominal
references. Adapted verifiers use LoRA \cite{hu2022lora} with rank 8,
$\alpha=16$, dropout 0.05, one training epoch, and learning rate
$2\times10^{-4}$. Training uses a per-GPU batch of two and four gradient accumulation steps. The field-loss weights in Eq.~\eqref{eq:loss} are
$(\lambda_y,\lambda_d,\lambda_e)=(10,3,1)$. Adapter training and MLLM evaluation use a maximum image area of 147,456 pixels. Evaluation uses batches of four, greedy decoding, and at most 180 new tokens, including for the unadapted backbones.

We use MVTec AD \cite{bergmann2019mvtec}, MVTec LOCO
\cite{bergmann2022loco}, VisA \cite{zou2022visa}, and KSDD2
\cite{bozic2021ksdd2}, with validation/test sizes of 732/1,090, 594/895, 1,299/2,016, and 402/602, respectively. For the first three datasets, adapter training uses 80\% of the official normal training images and 50\% of the official test anomalies, stratified by defect type. EfficientAD uses only the same 80\% nominal training subset and is scored on identical evaluation queries. Only adapter-training data are resampled: all normal images are retained, and anomalies are resampled with seed 42 to normal-to-abnormal ratios of 2:1 for MVTec AD and MVTec LOCO and 3:1 for VisA. Validation and test distributions are unchanged. For the three annotated benchmarks, Eq.~\eqref{eq:objective} uses
$a=1$, $b=0.5$, $c_{\mathrm r}=0.1$, $B=0.70$, and the shared grid
$\mathcal Q=\{0.03,0.04,\ldots,1.00\}$. Each dataset--backbone pair selects $q^\star$ on validation data only. All components and routing parameters are then frozen for test evaluation.

\begin{table}[t]
\centering
\caption{Average accuracy under different datasets.}
\label{tab:macro}
\setlength{\tabcolsep}{7pt}
\normalsize
\begin{tabular}{@{}l r r c@{}}
\toprule
Method & Acc. (\%) & BAcc. (\%) & Invocation rate (\%) \\
\midrule
\method{} & \textbf{93.20} & \textbf{86.83} & \textbf{45.08} \\
SAEC & 88.04 & 79.06 & 78.24 \\
MLLM & 79.95 & 77.24 & 100.00 \\
EfficientAD & 83.51 & 83.02 & 0.00 \\
\midrule
\method{} & \textbf{91.62} & \textbf{83.91} & \textbf{43.85} \\
SAEC & 86.20 & 76.16 & 78.24 \\
MLLM & 80.74 & 65.99 & 100.00 \\
EfficientAD & 83.51 & 83.02 & 0.00 \\
\bottomrule
\end{tabular}
\end{table}

\textbf{Baselines and Evaluation Metrics.}
We compare four baselines. \emph{EfficientAD} uses the local detector alone. The \emph{\unadapted{}} invokes the corresponding backbone on every image without task-specific adaptation. \emph{Controlled SAEC} applies the scene-complexity/local-screening route from SAEC \cite{tian2025saec} but uses the same three-field LoRA-adapted verifier as UACR. It is not the original
binary-adapter configuration. UACR applies Eq.~\eqref{eq:gate}.
All semantic paths share the query/reference protocol, prompt, output schema, and scoring rules.

For the true anomaly
indicator $y_i$, final abnormal prediction $\hat y_i$, and correct normalized
defect-type match $z_i$, the semantic scores are: $ \thitA=\frac{\sum_i y_i z_i}{\sum_i y_i}, \thitP=\frac{\sum_i y_i\hat y_i z_i}{\sum_i\hat y_i}$.
On type-annotated benchmarks, TypeHit-A measures type recall over true
anomalies. TypeHit-P measures successful type
identification among predicted anomalies, counting false-positive alarms in
its denominator. TypeF1 is their harmonic mean, $2AP/(A+P)$, where $A$ and
$P$ denote TypeHit-A and TypeHit-P, respectively. TypeF1 is zero when both are
zero. ExplainHit is the fraction of explanation-annotated anomalies whose
generated text matches the required evidence keywords. Invocation rate is the fraction of queries sent to the MLLM. The latency is end-to-end latency, measured from image input to the final structured inspection output, including local detection, routing, and MLLM verification when invoked.

\subsection{Main Results}

\textbf{Accuracy Comparison.}
Tables~\ref{tab:macro} and~\ref{tab:dataset-accuracy} report the accuracy results, with balanced accuracy additionally reported in Table~\ref{tab:macro}. In both tables, the upper and lower blocks present results for \qwenNine{} and \qwenSeven{}, respectively. Across the four datasets, UACR improves accuracy over SAEC by 5.16\% and 5.42\% for \qwenNine{} and \qwenSeven{}, respectively, and outperforms EfficientAD by 9.69 and 8.11 percentage points. This improvement is consistent across benchmarks rather than being driven by a single dataset: UACR achieves the highest accuracy for both backbones on MVTec AD, MVTec LOCO, and VisA. The largest gain is observed on MVTec AD, where UACR with \qwenNine{} reaches 97.8\%, compared with 87.9\% for controlled SAEC and 88.3\% for EfficientAD. On MVTec LOCO, UACR exceeds 84\% accuracy with both MLLM backbones.These results show that uncertainty-aware routing consistently improves accuracy across datasets and backbones, with the largest benefit on MVTec AD, where fine-grained anomaly discrimination is most demanding.

\begin{table}[th]
\centering
\caption{Binary accuracy under different datasets.}
\label{tab:dataset-accuracy}
\setlength{\tabcolsep}{3pt}
\normalsize
\renewcommand{\arraystretch}{1.12}
\begin{tabular*}{\columnwidth}{@{\extracolsep{\fill}}ccccc@{}}
\toprule
Dataset & \method{} & SAEC &
MLLM & EfficientAD \\
\midrule
MVTec AD   & \textbf{97.80} & 87.89 & 89.91 & 88.35 \\
MVTec LOCO & \textbf{84.36} & 82.46 & 50.84 & 76.76 \\
VisA       & \textbf{94.64} & 90.13 & 84.38 & 83.88 \\
KSDD2 & \textbf{96.01} & 91.69 & 94.68 & 85.05 \\
\midrule
MVTec AD   & \textbf{95.41} & 85.41 & 80.55 & 88.35 \\
MVTec LOCO & \textbf{84.25} & 81.45 & 68.27 & 76.76 \\
VisA       & \textbf{93.80} & 89.58 & 85.12 & 83.88 \\
KSDD2 & \textbf{93.02} & 88.37 & 89.04 & 85.05 \\
\bottomrule
\end{tabular*}
\end{table}

\begin{figure}[th]
  \centering
  \includegraphics[width=0.95\columnwidth]{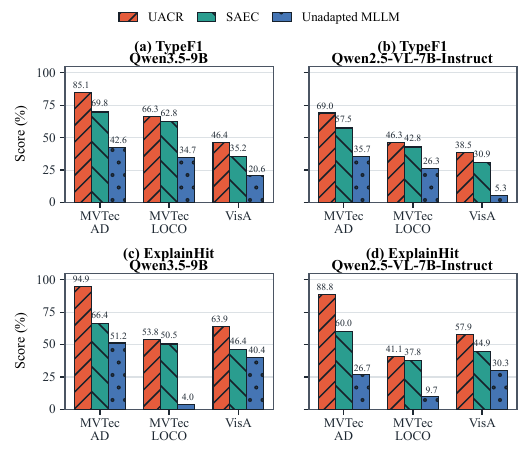}
    \caption{Defect-semantic interpretability comparison. }
  \label{fig:semantics}
\end{figure}

\textbf{Defect-Semantic Interpretability.}
Fig.~\ref{fig:semantics} shows that UACR achieves the highest TypeF1 across all six annotated dataset--backbone pairs. On MVTec AD, TypeF1 rises from 69.8\% for controlled SAEC to 85.1\% for UACR with \qwenNine{}, and from 57.5\% to 69.0\% with \qwenSeven{}; the corresponding UACR ExplainHit scores are 94.9\% and 88.8\%. On VisA, whose defect-type vocabulary is more diverse, UACR improves TypeF1 for \qwenNine{} by 11.2 percentage points and ExplainHit by 17.5 percentage points over controlled SAEC. Across both backbones, UACR improves macro TypeHit-P by at least 1.4\% over controlled SAEC and by at least 34.3\% over the \unadapted{}. These results indicate that uncertainty-aware routing not only improves anomaly discrimination but also yields defect descriptions that are more semantically faithful and better aligned with ground-truth defect types.

\begin{figure}[th]
  \centering
  \includegraphics[width=0.95\columnwidth]{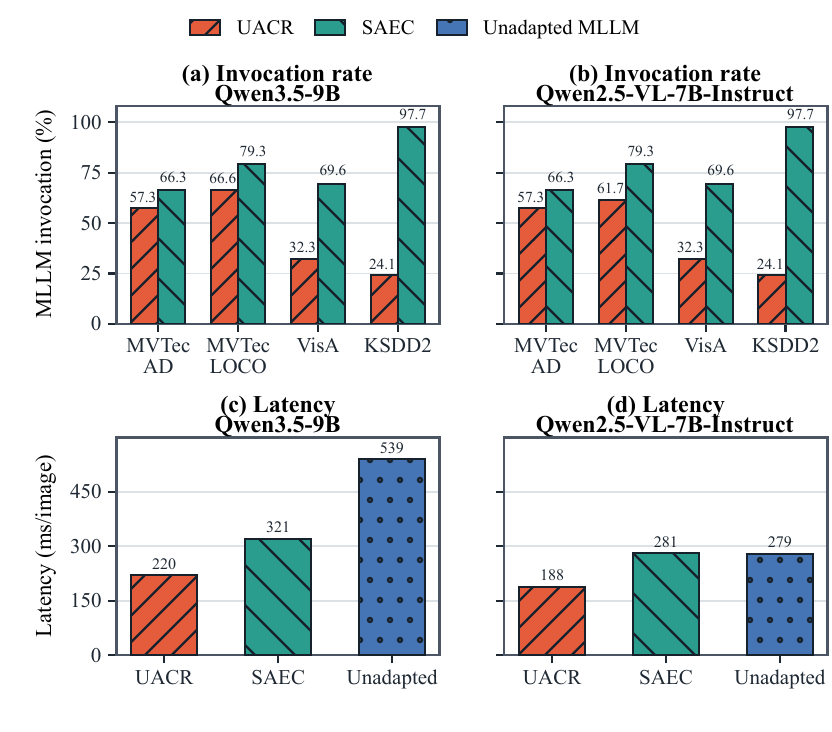}
  \caption{MLLM invocation rate and latency comparison under different methods.}
  \label{fig:invocation}
\end{figure}

\textbf{Latency Comparison.}
 Fig.~\ref{fig:invocation} shows that the reduction in invocations holds for every dataset--backbone pair. UACR invokes the 9B and 7B verifiers for 45.1\% and 43.9\% of queries, respectively, compared with 78.2\% for controlled SAEC and 100\% for the \unadapted{}. It attains higher macro accuracy with substantially fewer verifier calls. Across both backbones, UACR reduces latency by at least 31.5\% relative to controlled SAEC and by at least 32.6\% relative to the \unadapted{} backbones.  Overall, UACR improves the accuracy--latency tradeoff by concentrating computation on uncertain samples while bypassing the verifier for confident ones.

\section{Conclusion}
We presented UACR, an uncertainty-aware collaborative routing framework for explainable industrial anomaly inspection. UACR addresses the tension among detection accuracy, inference efficiency, and interpretability by estimating per-sample uncertainty and adaptively dispatching each input across inspection paths. Experiments on multiple public datasets show that UACR improves accuracy by over 5.16\% over the baseline while reducing latency by over 31.5\%, and enhances interpretability by over 15\%. These results indicate that UACR can deliver reliable anomaly decisions and operator-facing explanations under strict latency budgets.

\FloatBarrier
\raggedbottom
\begingroup
\normalsize
\bibliographystyle{IEEEbib}
\bibliography{references}
\endgroup

\end{document}